\documentclass[11pt]{article}

\usepackage[]{ACL2023}
\usepackage{times}
\usepackage{latexsym}

\usepackage{comment}
\usepackage{graphicx,xcolor} 
\usepackage{pxrubrica}        
\usepackage{url}
\usepackage{comment}
\usepackage{enumerate}
\usepackage{booktabs}
\usepackage{multirow}
\usepackage{amsmath}
\usepackage{amssymb}
\usepackage{amsfonts}
\usepackage{pifont}
\usepackage{caption}
\usepackage{subcaption}
\usepackage{here}
\usepackage{accents}
\usepackage{contour}
\usepackage{tabularx}
\usepackage[whole]{bxcjkjatype} 
\usepackage{wasysym}
\usepackage{pxrubrica}
\usepackage{enumitem}

\newcommand{\ctext}[1]{\raise0.2ex\hbox{\textcircled{\scriptsize{#1}}}}
\newcommand{\cmark}{\textcolor{green}{\ding{51}}}%
\newcommand{\xmark}{\textcolor{red}{\ding{55}}}%

\usepackage{tikz}
\usetikzlibrary{decorations.pathmorphing}
\newcommand{\myuwave}[1]{%
  \tikz[baseline=(char.base)]{
    \node[inner sep=0pt,outer sep=0pt] (char) {#1};
    \draw[red,decorate,decoration={snake,amplitude=0.6pt,segment length=3pt}] 
      ([yshift=-1.0pt]char.south west) -- ([yshift=-1.0pt]char.south east);
  }%
}

\usepackage{soul}
\setul{0.1ex}{0.1ex}

\usepackage[T1]{fontenc}

\usepackage[utf8]{inputenc}

\usepackage{microtype}

\usepackage{inconsolata}

\title{How do different tokenizers perform on downstream tasks \\in scriptio continua languages?: A case study in Japanese}

 \author{Takuro Fujii{\rm $^{\text{*}}$} \\ Yokohama National University \\  \texttt{tkr.fujii.ynu@gmail.com}
        \And Koki Shibata{\rm $^{\text{*}}$} \\ University of Tsukuba \\ \texttt{s1811496@klis.tsukuba.ac.jp}\\
        \AND
        Atsuki Yamaguchi{\rm ,} Terufumi Morishita \and Yasuhiro Sogawa \\ Hitachi,  Ltd. \\ \texttt{\{atsuki.yamaguchi.xn,terufumi.morishita.wp,yasuhiro.sogawa.tp\}@hitachi.com}}

\begin{document}
\maketitle

{
\let\thefootnote\relax\footnotetext{$^{\text{*}}$ Work done while interning at Hitachi, Ltd.}
}

\begin{abstract}
This paper investigates the effect of tokenizers on the downstream performance of pretrained language models (PLMs) in \textit{scriptio continua} languages where no explicit spaces exist between words, using Japanese as a case study.
The tokenizer for such languages often consists of a morphological analyzer and a subword tokenizer, requiring us to conduct a comprehensive study of all possible pairs.
However, previous studies lack this comprehensiveness.
We therefore train extensive sets of tokenizers, build a PLM using each, and measure the downstream performance on a wide range of tasks.
Our results demonstrate that each downstream task has a different optimal morphological analyzer, and that it is better to use Byte-Pair-Encoding or Unigram rather than WordPiece as a subword tokenizer, regardless of the type of task.
\end{abstract}

\section{Introduction} \label{sec: introduction}
Tokenization is the first key procedure in current natural language processing when inputting a target sentence to a pretrained language model (PLM).
It generally splits an input sequence into subword units, where a subword is a fraction of a word.
Previous efforts have proposed several subword-tokenization algorithms (hereafter, subword tokenizers), such as Byte-Pair-Encoding (BPE)~\cite{sennrich-etal-2016-neural}, WordPiece~\cite{6289079}, and Unigram~\cite{kudo-2018-subword}, and different PLMs use different subword tokenizers.\footnote{For example, BERT~\cite{devlin-etal-2019-bert} uses WordPiece, and GPT-3~\cite{https://doi.org/10.48550/arxiv.2005.14165} uses byte-level BPE.}

\begin{figure}[t]
\begin{center}
\includegraphics[width=\columnwidth]{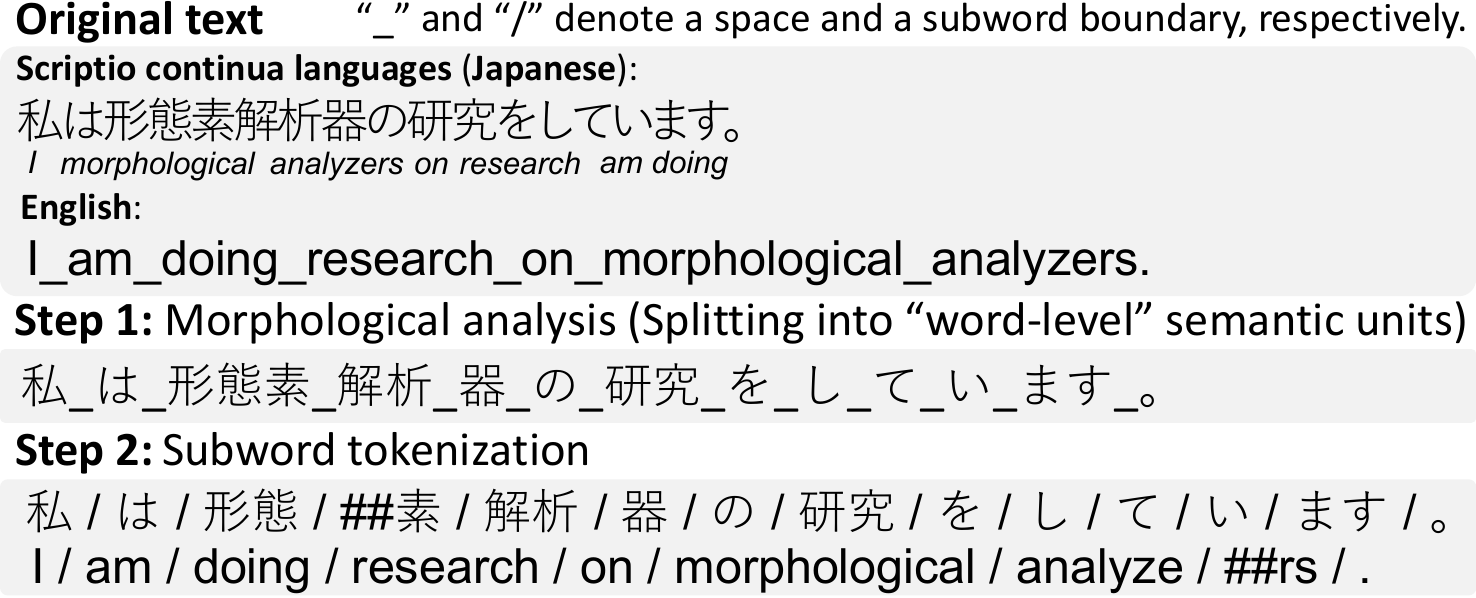}
\caption{Typical tokenization procedures in both \textit{scriptio continua} languages and English}
\label{figure:tokenization}
\end{center}
\end{figure}

It is widely acknowledged that tokenization affects the downstrem performance of PLMs~\cite{rust-etal-2021-good,gow-smith-etal-2022-improving,bostrom-durrett-2020-byte,park-etal-2020-empirical,https://doi.org/10.48550/arxiv.2204.08832}.
The majority of the previous studies have focused on languages with explicit word boundaries, such as English, while research on \textit{scriptio continua} languages, or languages without word boundaries (like Japanese, Chinese, and Thai), is still understudied.
The tokenization process in scriptio continua languages traditionally involves morphological analysis, which splits the input text into morphemes (semantic units similar to words in English) using the dictionary designed by human experts (see Step 1 in Figure \ref{figure:tokenization} for an example).
In this case, a tokenizer for a PLM consists of a morphological analyzer and a subword tokenizer.
To investigate the impact of tokenization in this scenario, we need to perform a comprehensive study on several sets of the available pairs, which is lacking in the previous work \cite{bostrom-durrett-2020-byte,Inoue2022GPT,DBLP:journals/corr/abs-2101-09635}.

In this paper, we investigate the effect of tokenizers on the downstream performance of PLMs in scriptio continua languages, focusing on Japanese as a case study.
We train an extensive collection of tokenizers consisting of known morphological analyzer and subword tokenizer pairs, use them to pretrain and fine-tune BERT models, and measure their performance on a variety of downstream tasks.
On the basis of the experimental results, we address the following three research questions.
We first try to answer if we should use a morphological analyzer\footnote{Not using a morphological analyzer means that we apply subword tokenization directly, the same as in cross-lingual PLMs such as XLM-R~\cite{conneau-etal-2020-unsupervised}.} in a scriptio continua language (Japanese) (RQ1).
RQ2 and RQ3 each examine whether different morphological analyzers/subword tokenizers perform differently on a downstream task.

\paragraph{Contributions}
1) We test a comprehensive set of known morphological analyzer and subword tokenizer pairs and use various downstream tasks to clarify the effect of tokenizers on the downstream performance of Japanese PLMs.
2) Accordingly, we find the followings:
\setlist{nolistsep}
\begin{itemize}[noitemsep]
    \item We should use a morphological analyzer for Japanese.
    \item Each task seems to have its own optimal morphological analyzer(s).
    \item It is better to use either BPE or Unigram as a subword tokenizer rather than WordPiece.
\end{itemize} 
3) We publicly release the code and PLMs.\footnote{Available at \url{https://github.com/hitachi-nlp/compare-ja-tokenizer}.}

\section{Japanese Tokenizer} \label{tokenizers}
In this section, we explain the morphological analyzers and subword tokenizers used in this paper.

\subsection{Japanese Morphological Analyzers}
Japanese morphological analyzers are based on either a pointwise or sequence prediction method.
The former tokenizes a sentence by extracting features from the characters within a pre-defined window and then predicting if a boundary exists between each character using a classifier.
The latter first constructs a lattice from an input sentence on the basis of a pre-defined dictionary, where each path in the lattice represents a candidate token sequence and has a cost, and then selects the path with the lowest cumulative cost as the analysis result.\footnote{Since it is intractable to compute costs for all candidate paths, previous studies have used either the Viterbi algorithm~\cite{1054010} or beam search to select a path.}
We obtain a cost for each path using a statistical model(s) or a hand-crafted dictionary.

We test the following four widely used morphological analyzers: MeCab \ctext{M}~\cite{kudo-etal-2004-applying}, Juman++ \ctext{J}~\cite{tolmachev-etal-2018-juman}, Sudachi \ctext{S}~\cite{takaoka-etal-2018-sudachi}, and Vaporetto \ctext{V}~\cite{vaporetto}. The first three adopt sequence prediction while the last uses pointwise prediction.\footnote{For more details, refer to Appendix \ref{appendix: morph}.}

\subsection{Subword Tokenizers} \label{sec:2.2}
We compare the following three tokenizers: BPE ($\mathcal{B}$), WordPiece ($\mathcal{W}$), and Unigram ($\mathcal{U}$), each of which differs in either vocabulary construction, tokenization algorithms, or both.
These tokenizers are empirically known to produce different subword boundaries~\cite{bostrom-durrett-2020-byte}.

\paragraph{Vocabulary Construction}
BPE constructs the vocabulary by merging and adding a pair of existing tokens with the highest score in the dictionary until the total number of tokens in the dictionary reaches a pre-defined size.
The score is calculated based on the frequency of the existing tokens.
WordPiece is similar to BPE but calculates the score based on the frequency of a symbol pair and the individual frequencies.
Unigram heuristically builds a large seed vocabulary from a training corpus (e.g., by taking the most frequent substrings) and then iteratively removes the least important symbols from the vocabulary.
Specifically, it first fits a unigram LM for the current vocabulary and then computes (i) the log likelihood of the training corpus with the LM and (ii) that of the training corpus with the LM after removing a particular symbol.
It then sets $\text{(i)}-\text{(ii)}$ as the cost, which shows the degradation of the log likelihood when the symbol is removed.
Finally, it removes the symbol with the lowest degradation.

\paragraph{Tokenization}
BPE splits a word into characters and iteratively merges those with the most frequent pair into larger known symbols in the vocabulary.
WordPiece\footnote{We follow the longest-match-first strategy used in BERT.} splits a word by the longest subword starting at the beginning of the word in the dictionary and continues splitting until its end.
Unigram tokenizes a word by performing Viterbi inference to select the maximum likelihood segmentation based on its vocabulary and unigram LM.

\renewcommand*{\arraystretch}{1.0}
\begin{table*}[t]
\begin{center}
\small
\resizebox{\linewidth}{!}{%
\begin{tabular}{llccccccc|c}
\toprule
\multicolumn{2}{l}{\textbf{Tokenizer}} & \textbf{MARC-ja} & \textbf{JSTS} & \textbf{JNLI} & \textbf{JSQuAD} & \textbf{JCQA} & \textbf{NER} & \multicolumn{1}{c|}{\textbf{UD}} & \textbf{Avg.} \\
\scriptsize Subword & \scriptsize Morphological & \scriptsize Accuracy & \scriptsize Spearman & \scriptsize Accuracy & \scriptsize F1 & \scriptsize Acc & \scriptsize F1 & \scriptsize LAS & \\

\midrule
\multicolumn{2}{l}{\texttt{bert-base-japanese}} & 95.5{\scriptsize$\pm$0.1} & 85.3{\scriptsize$\pm$0.3} & 86.8{\scriptsize$\pm$0.6} & 86.4{\scriptsize$\pm$0.2} & 76.6{\scriptsize$\pm$0.8} & 85.6{\scriptsize$\pm$0.2} & 93.3{\scriptsize$\pm$0.1} & 87.1 \\

\midrule
 & \ctext{M}~MeCab & $\myuwave{\text{95.4}}${\scriptsize$\pm$0.2} & 84.2{\scriptsize$\pm$0.1} & 88.0{\scriptsize$\pm$0.4} & 90.1{\scriptsize$\pm$0.3} & 74.1{\scriptsize$\pm$0.7} & 83.7{\scriptsize$\pm$0.8} & 93.6{\scriptsize$\pm$0.1} & 87.0 \\
& \ctext{J}~Juman++ & 95.5{\scriptsize$\pm$0.1} & 84.6{\scriptsize$\pm$0.4} & 87.6{\scriptsize$\pm$0.4}  & 90.1{\scriptsize$\pm$0.2} & 73.8{\scriptsize$\pm$0.3} & 85.1{\scriptsize$\pm$0.6} & 93.6{\scriptsize$\pm$0.1} & 87.2 \\
BPE& \ctext{S}~Sudachi & 95.5{\scriptsize$\pm$0.1} & 84.2{\scriptsize$\pm$0.2} & 88.2{\scriptsize$\pm$0.3} & 90.2{\scriptsize$\pm$0.2} & 74.2{\scriptsize$\pm$0.6} & 83.5{\scriptsize$\pm$0.6} & 93.8{\scriptsize$\pm$0.1} & 87.1 \\
($\mathcal{B}$)& \ctext{V}~Vaporetto & 95.6{\scriptsize$\pm$0.1} & 84.8{\scriptsize$\pm$0.2} & 87.5{\scriptsize$\pm$0.3} & 89.9{\scriptsize$\pm$0.2} & 74.2{\scriptsize$\pm$1.1} & 84.1{\scriptsize$\pm$0.9} & 93.7{\scriptsize$\pm$0.1} & 87.1 \\
& Nothing & $\myuwave{\text{95.4}}${\scriptsize$\pm$0.2} & $\myuwave{\text{82.8}}${\scriptsize$\pm$0.2} & $\myuwave{\text{87.2}}${\scriptsize$\pm$0.2} & $\myuwave{\text{88.7}}${\scriptsize$\pm$0.3} & $\myuwave{\text{72.8}}${\scriptsize$\pm$0.8} & $\myuwave{\text{62.9}}${\scriptsize$\pm$1.1} & $\myuwave{\text{93.4}}${\scriptsize$\pm$0.1} & $\myuwave{\text{83.3}}$ \\

\midrule
 & MeCab & 95.5{\scriptsize$\pm$0.1} & 82.4{\scriptsize$\pm$0.5} & 87.5{\scriptsize$\pm$0.3} & 89.2{\scriptsize$\pm$0.3} & 69.8{\scriptsize$\pm$0.7} & 84.0{\scriptsize$\pm$0.9} & 93.6{\scriptsize$\pm$0.1} & 86.0 \\
 & Juman++ & 95.3{\scriptsize$\pm$0.3} & 83.3{\scriptsize$\pm$0.3} & 87.7{\scriptsize$\pm$0.2}  & 89.8{\scriptsize$\pm$0.3} & 71.1{\scriptsize$\pm$0.6} & 84.7{\scriptsize$\pm$0.5} & 93.6{\scriptsize$\pm$0.1} & 86.5 \\
 WordPiece & Sudachi & 95.3{\scriptsize$\pm$0.2} & 83.7{\scriptsize$\pm$0.3} & 87.2{\scriptsize$\pm$0.4} & 89.6{\scriptsize$\pm$0.1} & 70.0{\scriptsize$\pm$0.9} & 82.4{\scriptsize$\pm$0.6} & 94.0{\scriptsize$\pm$0.1} & 86.0 \\
 ($\mathcal{W}$) & Vaporetto & 95.3{\scriptsize$\pm$0.2} & 83.6{\scriptsize$\pm$0.1} & 88.0{\scriptsize$\pm$0.4} & 89.7{\scriptsize$\pm$0.2} & 71.0{\scriptsize$\pm$0.4} & 84.0{\scriptsize$\pm$0.8} & 93.8{\scriptsize$\pm$0.1} & 86.5 \\
 & Nothing & $\myuwave{\text{85.5}}${\scriptsize$\pm$0.0} & N/A & $\myuwave{\text{55.3}}${\scriptsize$\pm$0.0} & $\myuwave{\text{10.1}}${\scriptsize$\pm$0.1} & $\myuwave{\text{20.0}}${\scriptsize$\pm$0.8} & $\myuwave{\text{0.0}}${\scriptsize$\pm$0.0} & $\myuwave{\text{63.8}}${\scriptsize$\pm$0.9} & $\myuwave{\text{33.5}}$ \\

 \midrule
 & MeCab & $\myuwave{\text{95.4}}${\scriptsize$\pm$0.3} & 84.6{\scriptsize$\pm$0.4} & 88.3{\scriptsize$\pm$0.4} & 89.5{\scriptsize$\pm$0.3} & 74.5{\scriptsize$\pm$0.8} & 83.1{\scriptsize$\pm$1.0} & 93.4{\scriptsize$\pm$0.2} & 87.0 \\
 & Juman++ & $\myuwave{\text{95.4}}${\scriptsize$\pm$0.2} & 84.3{\scriptsize$\pm$0.3} & 87.8{\scriptsize$\pm$0.3} & 89.9{\scriptsize$\pm$0.2} & 74.9{\scriptsize$\pm$1.2} & 84.1{\scriptsize$\pm$0.4} & 93.4{\scriptsize$\pm$0.1} & 87.1 \\
 Unigram & Sudachi & 95.6{\scriptsize$\pm$0.2} & 84.8{\scriptsize$\pm$0.5} & 88.4{\scriptsize$\pm$0.3} & 89.9{\scriptsize$\pm$0.1} & 74.5{\scriptsize$\pm$0.6} & 83.0{\scriptsize$\pm$1.3} & 93.7{\scriptsize$\pm$0.1} & 87.1 \\
 ($\mathcal{U}$) & Vaporetto & 95.5{\scriptsize$\pm$0.3} & 84.6{\scriptsize$\pm$0.2} & 87.9{\scriptsize$\pm$0.3} & 89.9{\scriptsize$\pm$0.1} & $\myuwave{\text{74.3}}${\scriptsize$\pm$0.8} & 84.1{\scriptsize$\pm$0.4} & 93.7{\scriptsize$\pm$0.1} & 87.1 \\
 & Nothing & $\myuwave{\text{95.4}}${\scriptsize$\pm$0.4} & $\myuwave{\text{83.9}}${\scriptsize$\pm$0.3} & $\myuwave{\text{87.7}}${\scriptsize$\pm$0.8} & $\myuwave{\text{89.3}}${\scriptsize$\pm$0.1} & 74.6{\scriptsize$\pm$0.4} & $\myuwave{\text{76.9}}${\scriptsize$\pm$1.0} & $\myuwave{\text{93.2}}${\scriptsize$\pm$0.2} & $\myuwave{\text{85.9}}$ \\
\bottomrule
\toprule
\multicolumn{10}{l}{\textbf{Statistical test results}: Kruskal-Wallis test \cite{doi:10.1080/01621459.1952.10483441}. \cmark~if $p <$ .05 otherwise \xmark.}\\
\multicolumn{2}{l}{RQ2: ($\mathcal{B}$, $\mathcal{W}$, $\mathcal{U}$)} & ({\scriptsize\xmark,~\xmark,~\xmark}) & ({\scriptsize\cmark,~\cmark,~\xmark}) & ({\scriptsize\cmark,~\xmark,~\xmark}) & ({\scriptsize\xmark,~\xmark,~\xmark}) & ({\scriptsize\xmark,~\cmark,~\xmark}) & ({\scriptsize\cmark,~\cmark,~\xmark}) & \multicolumn{1}{c}{({\scriptsize\cmark,~\cmark,~\cmark})}\\
\multicolumn{2}{l}{RQ3: (\ctext{M}, \ctext{J}, \ctext{S}, \ctext{V})} & ({\scriptsize\xmark,~\xmark,~\xmark,~\xmark}) & ({\scriptsize\cmark,~\cmark,~\cmark,~\cmark}) & ({\scriptsize\xmark,~\xmark,~\cmark,~\xmark}) & ({\scriptsize\cmark,~\xmark,~\cmark,~\xmark}) & ({\scriptsize\cmark,~\cmark,~\cmark,~\cmark}) & ({\scriptsize\xmark,~\xmark,~\xmark,~\xmark}) & \multicolumn{1}{c}{({\scriptsize\xmark,~\xmark,~\cmark,~\xmark})} &\\
\bottomrule
\end{tabular}%
}
\caption{Results from seven tasks with standard deviations over five runs.
JCQA stands for JCommonsenseQA.
$\myuwave{\text{Values with a wavy line}}$ denote the worst results among morphological analyzers with the same subword tokenizer.
\cmark~indicates that there is statistical significance among (RQ2) morphological analyzers with the same subword tokenizer or (RQ3) subword tokenizers with the same morphological analyzer, while \xmark~denotes that there is no statistical significance. For example, ({\scriptsize\cmark,~\xmark,~\xmark}) in RQ2 indicates that there is statistical significance between different morphological analyzers with BPE, while no statistical significance is observed for WordPiece or Unigram.}
\label{table:result}
\end{center}
\end{table*}

\section{Experimental Setup\footnote{For implementation details, refer to Appendix \ref{appendix:impl}.}}

\paragraph{Tokenizers}
We compared a total of 12 tokenizers (four morphological analyzers and three subword tokenizers), as introduced in \S\ref{tokenizers}.
We also considered three additional tokenizers not using morphological analyzers.
We trained all tokenizers with the vocabulary size of 30k utilizing 10M sentences randomly extracted from Japanese Wikipedia.

\paragraph{Models}
We used the base configuration of BERT (total parameters: 125M).
For each tokenizer, we pretrained BERT for 500k steps with masked language modeling \cite{devlin-etal-2019-bert} on the Japanese Wikipedia and CC-100~\cite{conneau-etal-2020-unsupervised} datasets, consisting of 2.2 and 1.1M samples each with the maximum length set to 512.

\paragraph{Benchmarks}
We used the following benchmarks: JGLUE~\cite{kurihara-etal-2022-jglue}, NER\footnote{\href{https://github.com/stockmarkteam/ner-wikipedia-dataset}{Dataset: \nolinkurl{stockmarkteam/ner-wikipedia-dataset}}}, and Universal Dependencies (UD) Japanese-GSD~\cite{asahara-etal-2018-universal}.\footnote{We provide the description of each task in Appendix \ref{appendix:task}.
For reference, we also measured the performance of \href{https://huggingface.co/cl-tohoku/bert-base-japanese}{\nolinkurl{bert-base-japanese}}, which uses MeCab and WordPiece.}
Since the test set for JGLUE is not publicly available, we fine-tuned all models on the training set using five-fold cross-validation and evaluated their performance on the development set.
Since the development and test sets are not available for NER, we split the training set into 9:1. We fine-tuned the models with five-fold cross-validation by the former and measured the performance using the latter.

\section{Results and Analysis} \label{sec: results}
This section addresses the three RQs raised in \S\ref{sec: introduction}.

\paragraph{RQ1: Should we use a morphological analyzer?}
Table \ref{table:result} lists the results on the seven downstream tasks grouped by subword tokenizer.
The average scores across tasks (``Avg.'') show that tokenizers without a morphological analyzer (``Nothing'') exhibited the worst results among tokenizers with the same subword tokenizer.
This trend also generally holds for task-specific results.
These results make intuitive sense because a morphological analyzer can provide explicit semantic boundaries of an input text, making the input units for subword tokenization similar to English words (Figure \ref{figure:tokenization}).
This should help a model to capture the semantic and syntactic information more easily and consequently outperform those that do not use a morphological analyzer.
We therefore conclude that we should use a morphological analyzer for Japanese.

In addition to the above, we observe that WordPiece + Nothing produced by far the worst results in all tasks due to the poor tokenization.
WordPiece processes a sequence word by word and treats a sequence without a blank as a single word.
If it fails to tokenize a particular word, it tokenizes the ``whole'' as a single \texttt{[UNK]} token.
Without a morphological analyzer, the length of a word becomes abnormally long, making WordPiece more likely to produce an \texttt{[UNK]} token.
This means that the majority of an input text will be converted into \texttt{[UNK]} tokens, thus losing almost all of the content in the text.
In fact, the average sequence length and ratio of \texttt{[UNK]} per sample in pretraining were $1.15\pm3.28$ and $99.8\pm4.9$\%, respectively.
These caused unstable pretraining (see Appendix \ref{appendix: loss}).

Compared with other tasks, Nothing in NER showed a considerable performance degradation with a maximum difference of 22.2 (Juman++ vs. Nothing in BPE).
In NER, annotations are word-level and tend to align well with morphemes.
Since tokenizers with morphological analyzers split a morpheme into subword tokens, they can produce more linguistically motivated subword segmentation than Nothing, thus giving them an advantage.

\paragraph{RQ2: Do different morphological analyzers perform differently on downstream tasks?}
Looking at the statistical test results for RQ2 in Table \ref{table:result}\footnote{Note that we omit Nothing from the following analyses.}, we can see that there were significant performance differences between different morphological analyzers with the same subword tokenizers in some tasks, e.g., JSTS, NER, and UD.
In other words, different morphological analyzers could perform differently on different downstream tasks.

For tasks with statistical significance, we further ran the Steel-Dwass test~\cite{doi:10.1080/03610929108830487} to see which morphological analyzer had a significant performance difference from the others (Table \ref{table:result_stat_morph}).
We can observe task-specific trends for an effective morphological analyzer(s).
Specifically, for JSTS, Vaporetto performed well.
For NER, Juman++ was effective.
For UD, Sudachi performed well.
Therefore, each task seems to have its own optimal morphological analyzer(s).

\renewcommand*{\arraystretch}{1.0}
\begin{table}[t]
\small
\resizebox{\linewidth}{!}{%
\begin{tabular}{@{}lccccc@{}}
\toprule
         & \textbf{JSTS}                                                                     & \textbf{JNLI} & \textbf{JCQA} & \textbf{NER}       & \textbf{UD}                                                                                                               \\ 
         \midrule
BPE       & \begin{tabular}[c]{@{}l@{}}(\ctext{V} > \ctext{M})\\ (\ctext{V} > \ctext{S})\end{tabular} & --             & --           & (\ctext{J} > \ctext{S}) & \begin{tabular}[c]{@{}l@{}}(\ctext{S} > \ctext{M})\\ (\ctext{S} > \ctext{J})\end{tabular}                                             \\
\midrule
WordPiece & \begin{tabular}[c]{@{}l@{}}(\ctext{S} > \ctext{M})\\ (\ctext{V} > \ctext{M})\end{tabular}     & --           & --             & (\ctext{J} > \ctext{S}) & \begin{tabular}[c]{@{}l@{}}
    (\ctext{S} > \ctext{M})\\ (\ctext{S} > \ctext{J})\\ (\ctext{V} > \ctext{M})\\ (\ctext{V} > \ctext{J})
\end{tabular} \\
\midrule

Unigram   & --                                                                               & --           & --           & --                & --                                                                                                                         \\ 
\bottomrule
\end{tabular}
}
\caption{Combinations of morphological analyzers with statistical significance ($p <$ .05, Steel-Dwass test). ``--'' indicates no statistical significance observed. ``\ctext{A} > \ctext{B}'' indicates that morphological analyzer \ctext{A} is significantly better than morphological analyzer \ctext{B}.}
\label{table:result_stat_morph}
\end{table}

\begin{table*}[t]
\small
\begin{center}
\begin{tabular}{@{}lccccccc@{}}
\toprule
\textbf{} & \textbf{MARC-ja} & \textbf{JSTS}                                                                   & \textbf{JNLI}        & \textbf{JSQuAD}                                                                 & \textbf{JCQA}                                                                   & \textbf{NER} & \textbf{UD}         \\ \midrule
MeCab     & --               & \begin{tabular}[c]{@{}c@{}}($\mathcal{B}$ > $\mathcal{W}$)\\ ($\mathcal{U}$ > $\mathcal{W}$)\end{tabular} & --                   & ($\mathcal{B}$ > $\mathcal{W}$)                                                                & \begin{tabular}[c]{@{}c@{}}($\mathcal{B}$ > $\mathcal{W}$)\\ ($\mathcal{U}$ > $\mathcal{W}$)\end{tabular} & --           & --                  \\
\midrule
Juman++   & --               & \begin{tabular}[c]{@{}c@{}}($\mathcal{B}$ > $\mathcal{W}$)\\ ($\mathcal{U}$ > $\mathcal{W}$)\end{tabular} & --                   & --                                                                              & \begin{tabular}[c]{@{}c@{}}($\mathcal{B}$ > $\mathcal{W}$)\\ ($\mathcal{U}$ > $\mathcal{W}$)\end{tabular} & --           & --                  \\
\midrule
Sudachi   & --               & ($\mathcal{U}$ > $\mathcal{W}$)                                                            & ($\mathcal{U}$ > $\mathcal{W}$) & \begin{tabular}[c]{@{}c@{}}($\mathcal{B}$ > $\mathcal{W}$)\\ ($\mathcal{U}$ > $\mathcal{W}$)\end{tabular} & \begin{tabular}[c]{@{}c@{}}($\mathcal{B}$ > $\mathcal{W}$)\\ ($\mathcal{U}$ > $\mathcal{W}$)\end{tabular} & --           & ($\mathcal{U}$ > $\mathcal{W}$) \\
\midrule
Vaporetto & --               & \begin{tabular}[c]{@{}c@{}}($\mathcal{B}$ > $\mathcal{W}$)\\ ($\mathcal{U}$ > $\mathcal{W}$)\end{tabular} & --                   & --                                                                              & \begin{tabular}[c]{@{}c@{}}($\mathcal{B}$ > $\mathcal{W}$)\\ ($\mathcal{U}$ > $\mathcal{W}$)\end{tabular} & --           & --                  \\ 
\bottomrule
\end{tabular}
\end{center}
\caption{Combinations of subword tokenizers with statistical significance ($p <$ .05, Steel-Dwass test). ``--'' indicates no statistical significance observed. ``$\mathcal{X}$ > $\mathcal{Y}$'' indicates that subword tokenizer $\mathcal{X}$ is significantly better than subword tokenizer $\mathcal{Y}$.}
\label{table:result_stat_subword}
\end{table*}

\begin{figure}[t]
\begin{center}
\includegraphics[width=\columnwidth]{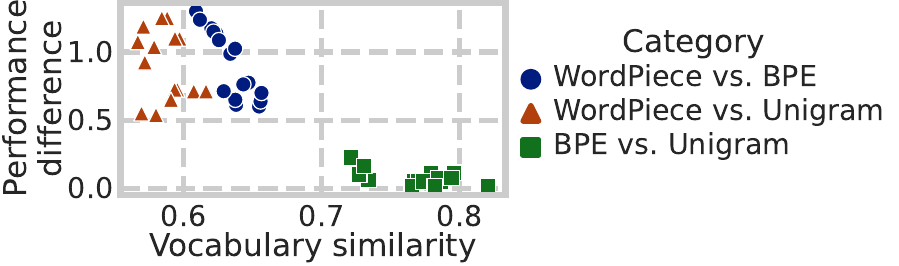}
\caption{Relationship between vocabulary similarity of subword tokenizers and their performance difference. Samples with the same subword tokenizer are excluded.}
\label{figure:correlation}
\end{center}
\end{figure}

\paragraph{RQ3: Do different subword tokenizers perform differently on downstream tasks?}
From the statistical test results for RQ3 in Table \ref{table:result}, we observe significant performance differences between subword tokenizers with the same morphological analyzers in some tasks, such as JSTS and JCQA.
``Avg.'' in Table \ref{table:result} indicates that WordPiece performed poorly, while BPE and Unigram achieved similar results.
The results of the Steel-Dwass test (Table \ref{table:result_stat_subword}) also confirmed that WordPiece showed significant performance degradation compared with either BPE, Unigram, or both in some tasks.
We did not observe a significant difference between BPE and Unigram across all tasks.
Therefore, different subword tokenizers could perform on downstream tasks differently, and it is better to use either BPE or Unigram.

We next analyze and discuss which differences in subword tokenizers produced downstream performance differences.
%
%
First, we look at the difference in the vocabulary of subword tokenizers.
We plot the relationship between vocabulary similarity and performance difference between two different subword tokenizers in Figure \ref{figure:correlation}.
The vocabulary similarity of two different subword tokenizers is computed as $\frac{|V_1 \cap V_2|}{|V|}$, where $|V|$ is the vocabulary size and $V_1$ and $V_2$ are the vocabularies of two subword tokenizers ($T_1$ and $T_2$).
For each task, we computed the performance difference between the two as $\frac{1}{5}|\sum_i s_{1i} - \sum_j s_{2j}|$, where $s_{1i}$ and $s_{2j}$ are the $i$-th and $j$-th observed scores of $T_1$ and $T_2$, respectively.
We observe that symbols related to WordPiece (\textcolor[HTML]{08217D}{\CIRCLE} and \textcolor[HTML]{AE431B}{$\blacktriangle$}) are plotted in the upper-left corner, while others (\textcolor[HTML]{1A6F22}{$\blacksquare$}) are in the lower-right corner, indicating that WordPiece has a different vocabulary composition than BPE and Unigram, and its performance difference is far larger than that between BPE and Unigram.
These results are consistent with our finding that WordPiece performed poorly with statistical significance, and both BPE and Unigram showed similar results.
Therefore, it is possible that the vocabulary of a subword tokenizer has something to do with the downstream performance.

%
%
Further, while WordPiece uses a greedy longest-match-first strategy in tokenizing a word, both BPE and Unigram use a more sophisticated approach (as explained in \S\ref{sec:2.2}). 
This algorithmic difference might also contribute to the performance difference between different subword tokenizers.

\section{Conclusion}
To investigate the effect of tokenizers on the downstream performance of PLMs in a scriptio continua language (Japanese), we compared extensive sets of tokenizers by evaluating them on a wide range of downstream tasks and addressed the three RQs in \S\ref{sec: introduction}.
Future work will examine how to automatically select the optimal tokenizer pair for a given task.

\section*{Limitations}
This study has the following limitations:
\begin{itemize}
    \item We fixed the vocabulary size of each subword tokenizer to 30k.
    Using a different size might yield different results than those in our paper, though the effect of varying the vocabulary size for a subword tokenizer seemed to be small if the size is sufficiently large (e.g., over 16k or more)~\cite{https://doi.org/10.48550/arxiv.2204.08832}.

    \item We have used the BERT architecture for our comparison, while there are other commonly used model architectures such as T5~\cite{10.5555/3455716.3455856} and GPT-3.
    The investigation with these architectures is our future work.

    \item To investigate the impact of tokenizers on the downstream performance of PLMs in scriptio continua languages, we have taken Japanese as a case study.
    Other scriptio continua languages will be addressed in the future.    
\end{itemize}

\section*{Ethics Statement}
This study did not involve any sensitive data but only used publicly available data, including Wikipedia, CC-100, JGLUE, Japanese NER, and UD as explained in the paper.
Although we plan to release the resulting models, they might perform unfairly in some circumstances, as reported in \citet{baldini-etal-2022-fairness}.
We highly recommend users to refer to studies on debiasing PLMs, such as \citet{guo-etal-2022-auto}.

\section*{Acknowledgements}
We would like to thank anonymous reviewers, Yuta Koreeda, and Yuichi Sasazawa for their insightful comments.
We also would like to thank Dr. Masaaki Shimizu for the maintenance and management of the large computational resources used in this paper.

\bibliography{anthology,custom}
\bibliographystyle{acl_natbib}

\clearpage
\appendix
\begin{table*}[t!]
\begin{center}
\small
\begin{tabular}{llcllll}
\toprule
\multicolumn{2}{l}{\multirow{2}{*}{\textbf{Dataset}}} & \multirow{2}{*}{\textbf{License}} & \multirow{2}{*}{\textbf{Task Type}} & \multicolumn{3}{c}{\textbf{Number of samples}} \\
\multicolumn{2}{l}{}                                  &                                   &                                     & Train           & Dev           & Test         \\ \midrule
\multirow{5}{*}{JGLUE}        & MARC-ja               & \multirow{5}{*}{CC BY-SA 4.0}     & Text classification                 & 187,528         & 5,654         & -            \\
                              & JSTS                  &                                   & Sentence pair classification        & 12,451          & 1,457         & -            \\
                              & JNLI                  &                                   & Sentence pair classification        & 20,073          & 2,434         & -            \\
                              & JSQuAD                &                                   & Question answering                  & 62,859          & 4,442         & -            \\
                              & JCommonsenseQA        &                                   & Question answering                  & 8,939           & 1,119         & -            \\ \midrule
\multicolumn{2}{l}{Japanese NER}          & CC-BY-SA 3.0                      & Named entity recognition            & 5,343           & -             & -            \\
\multicolumn{2}{l}{UD-Japanese-GSD}            & CC BY-SA 4.0                      & Dependency parsing                  & 7,050           & 507           & 543          \\ \bottomrule

\end{tabular}
\caption{Statistics for each dataset used in this paper. Note that the test sets are not currently publicly available for JGLUE. Japanese NER does not have the corresponding development and test sets.}
\label{table:statistics}
\end{center}
\end{table*}

\section*{Appendices}

\section{Japanese Morphological Analyzers} \label{appendix: morph}
\paragraph{MeCab~{\rm \cite{kudo-etal-2004-applying}}}
MeCab tokenizes a sentence by first constructing a lattice on the basis of its dictionary and then selecting the combination with the lowest cumulative cost using the Viterbi algorithm~\cite{1054010}.
The cost is calculated using a pre-defined feature function in sequence labeling.

\paragraph{Juman++~{\rm \cite{tolmachev-etal-2018-juman}}}
Juman++ tokenizes a sentence by constructing a lattice in accordance with the dictionary and subsequently selecting the path with the highest score by beam search. The score is calculated using both a RNN-based language model and a feature-based linear model.

\paragraph{Sudachi~{\rm \cite{takaoka-etal-2018-sudachi}}}
Sudachi puts an emphasis on offering a tokenizer and dictionary for business use, enabling us to select tokens of different granularity for each application. We use the ``Middle'' unit of granularity, which is similar to words in general sense.

\paragraph{Vaporetto~{\rm \cite{vaporetto}}}
Vaporetto tokenizes a sentence by extracting features from the characters within a pre-defined window and subsequently classifying if a boundary exists between each character with a linear classification model.

\section{Downstream Tasks} \label{appendix:task}
We briefly describe the seven downstream tasks used in this paper. The statistics for each task dataset are presented in Table \ref{table:statistics}.

\paragraph{MARC-ja}
A binary classification task to predict whether a product review is positive or negative.
The dataset is based on the Japanese part of the Multilingual Amazon Reviews Corpus (MARC)~\cite{keung-etal-2020-multilingual}.

\paragraph{JSTS}
A regression task to predict a semantic similarity score between two sentences.
The score ranges from 0 (least similar) to 5 (most similar).
The data were sourced from the Japanese version of the MS COCO Caption Dataset~\cite{https://doi.org/10.48550/arxiv.1504.00325} and the YJ Captions Dataset~\cite{miyazaki-shimizu-2016-cross}.

\paragraph{JNLI}
A three-way classification task to predict an inference relation between two sentences. The relation includes ``contradiction,'' ``neutral,'' and ``entailment,'' the same as in SNLI~\cite{bowman-etal-2015-large}.
The data source was the same as that for JSTS.

\paragraph{JSQuAD}
A question answering task to predict a corresponding answer span given a question and context.
The data were sourced from Japanese articles in Wikipedia and its construction process is based on SQuAD v1.1~\cite{rajpurkar-etal-2016-squad}.

\paragraph{JCommonsenseQA}
A multiple-choice question answering task to select the best choice from five choices given a question.
JCommonsenseQA is a Japanese version of CommonsenseQA~\cite{talmor-etal-2019-commonsenseqa}, and it was constructed in the same manner as in CommonsenseQA, which used the multilingual knowledge base: ConceptNet~\cite{Speer_Chin_Havasi_2017} as seeds.

\paragraph{NER}
A task to identify and categorize named entities in a given sentence. The data were sourced from Japanese articles in Wikipedia and annotated by Stockmark Inc. The dataset is available at \url{https://github.com/stockmarkteam/ner-wikipedia-dataset}.

\paragraph{UD}
A dependency parsing task to predict the syntactic dependency structure of a given sentence~\cite{zeman-etal-2017-conll,zeman-etal-2018-conll}.
The output is a directed tree originating out of a root node.
Each edge in the tree has a label that defines a grammatical relationship between two words.

\section{Implementation Details} \label{appendix:impl}
We implemented our tokenizers with the Tokenizers library\footnote{\url{https://github.com/huggingface/tokenizers}} and our models using the PyTorch \cite{NEURIPS2019_9015} and Transformers \cite{wolf-etal-2020-transformers} libraries. We trained our models with four NVIDIA V100 (32GB) GPUs for pretraining and one for fine-tuning. We used automatic mixed precision (FP16) provided by PyTorch as default.
The code is available on the GitHub: \url{https://github.com/hitachi-nlp/compare-ja-tokenizer}, and the models are available on the Hugging Face Hub: \url{https://huggingface.co/hitachi-nlp}.

\subsection{Data}
We downloaded Wikipedia data from \url{https://www.tensorflow.org/datasets/catalog/wikipedia\#wikipedia20201201ja}.
As its preprocessing step, we excluded sentences with less than 30 characters and those containing “\texttt{Category}” or table symbols.

\subsection{Model}
We used the base configuration of BERT (12 hidden layers and attention heads, Dim\textsubscript{hidden} $=$ 768, Dim\textsubscript{intermediate} $=$ 3072, Total parameters $=$ 125M).

\subsection{Pretraining}
We pretrained all models for 500k steps and optimized them with AdamW~\cite{loshchilov2018decoupled}.
We mostly followed the configurations of \citet{devlin-etal-2019-bert}.
Table \ref{table:hyperparams_pretraining} lists the hyperparameter settings used in pretraining.

\begin{table}[t]
\begin{center}
\small
\begin{tabular}{lc}
\toprule
\textbf{Hyperparameter} & \textbf{Value} \\
\midrule
Batch size & 128 \\
Total training steps & 500,000 \\
Adam $\epsilon$ & 1e-8 \\
Adam $\beta_1$ & 0.9 \\
Adam $\beta_2$ & 0.999 \\
Sequence length & 512 \\
Learning rate & 1e-4\\
Learning rate schedule & Linear warmup \\
Warmup steps & 10,000 \\
Weight decay & 0.01 \\
Attention dropout & 0.1 \\
Dropout & 0.1 \\
\bottomrule
\end{tabular}
\caption{Hyperparameters for pretraining}
\label{table:hyperparams_pretraining}
\end{center}
\end{table}

\renewcommand*{\arraystretch}{1.0}
\begin{table}[t]
\begin{center}
\small
\resizebox{\linewidth}{!}{%
\begin{tabular}{lc}
\toprule
\textbf{Hyperparameter} & \textbf{Value} \\
\midrule
Batch size & 32 \\
Epochs & 5 for JGLUE tasks \& NER \\
& 10 for UD\\
Adam $\epsilon$ & 1e-8 \\
Adam $\beta_1$ & 0.9 \\
Adam $\beta_2$ & 0.999 \\
Sequence length & 512 for MARC-ja \& UD\\
& 348 for JSQuAD\\
& 128 for JSTS, JNLI \& NER\\
& 64 for JCQA\\
Learning rate & 3e-5 for JGLUE tasks \& NER\\
& 5e-5 for BERT in UD\\
& 1e-3 for BAP in UD\\
Learning rate schedule & Linear warmup \\
Warmup steps & 10\% of steps\\
Weight decay & 0.01 \\
Attention dropout & 0.1 \\
Dropout & 0.1 \\
\bottomrule
\end{tabular}%
}
\caption{Hyperparameters for fine-tuning}
\label{table:hyperparams_finetuning}
\end{center}
\end{table}

\subsection{Fine-tuning}
Table \ref{table:hyperparams_finetuning} lists the hyperparameters for fine-tuning models on the JGLUE, NER, and UD datasets.
For UD, we trained a deep biaffine attention parser \cite{dozat2017deep} built on top of the PLMs.
We computed an average for each token over the top four layers of the BERT hidden representations and used it as an input to a biaffine attention parser (BAP). The dimensionalities of arc and relation features given to each biaffine module are 500 and 100, respectively.
We used the SuPar library\footnote{\url{https://github.com/yzhangcs/parser}} to implement the parser and followed its default hyperparameter configurations.

\section{Pretraining Loss} \label{appendix: loss}
\begin{figure*}[h]
\centering
\begin{minipage}[b]{0.45\textwidth}
    \centering
    \includegraphics[width=\textwidth]{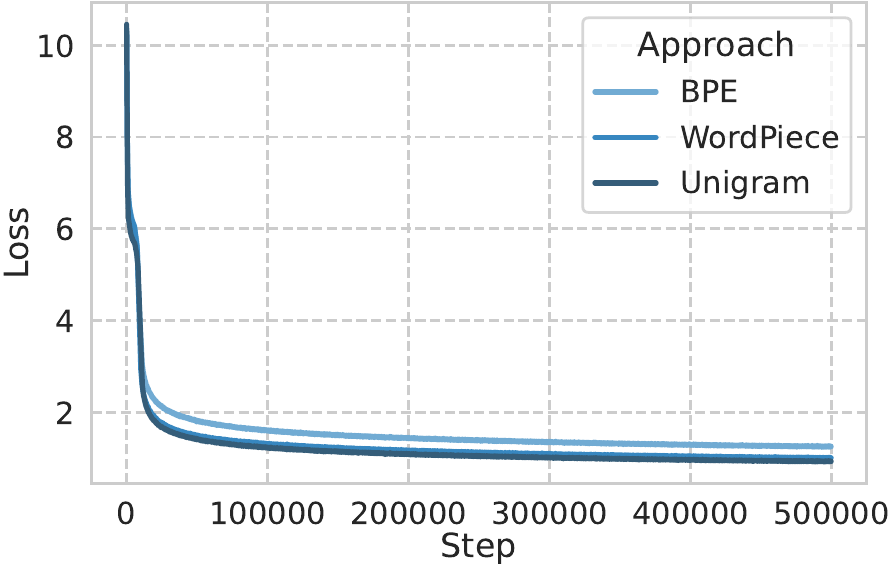}
    \subcaption{MeCab}
\end{minipage}
\hfill%
\begin{minipage}[b]{0.45\textwidth}
    \centering
    \includegraphics[width=\textwidth]{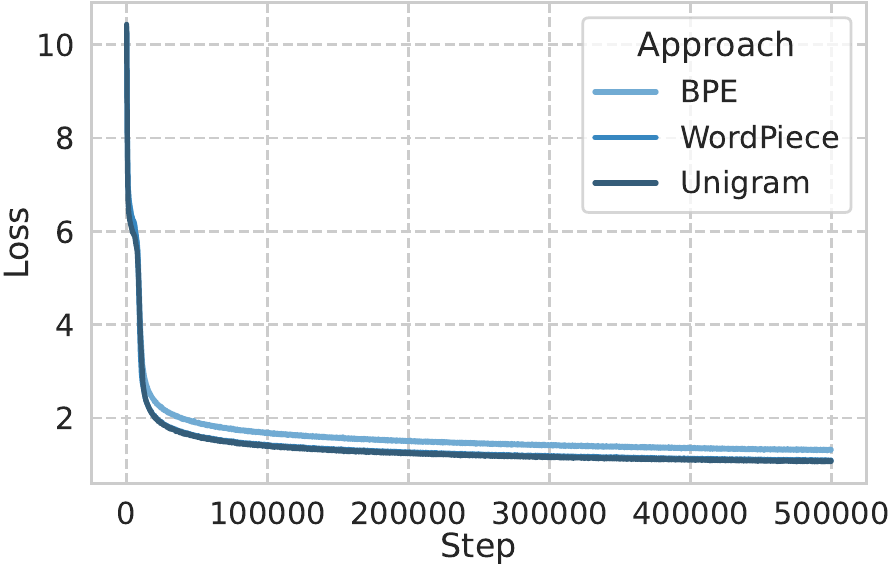}
    \subcaption{Juman++}
\end{minipage}
\hfill%
\begin{minipage}[b]{0.45\textwidth}
    \centering
    \includegraphics[width=\textwidth]{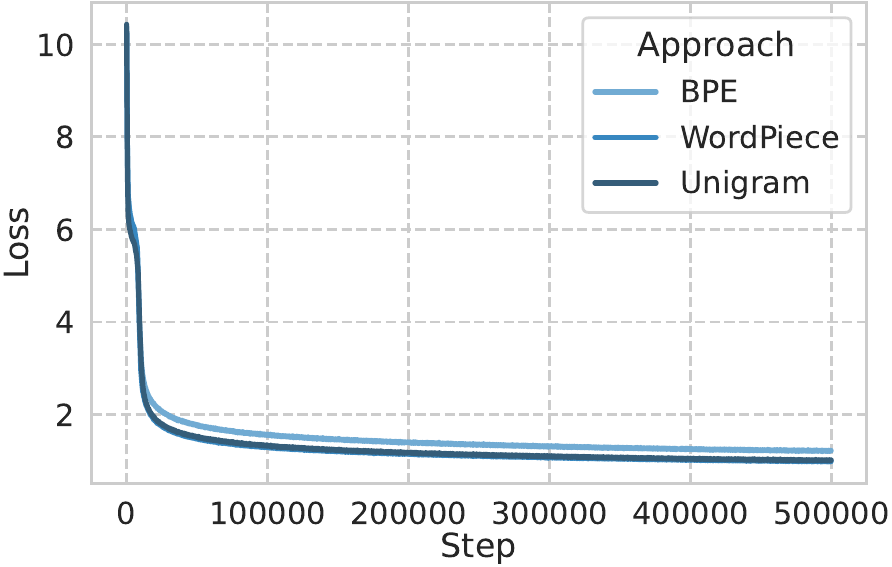}
    \subcaption{Sudachi}
\end{minipage}
\hfill%
\begin{minipage}[b]{0.45\textwidth}
    \centering
    \includegraphics[width=\textwidth]{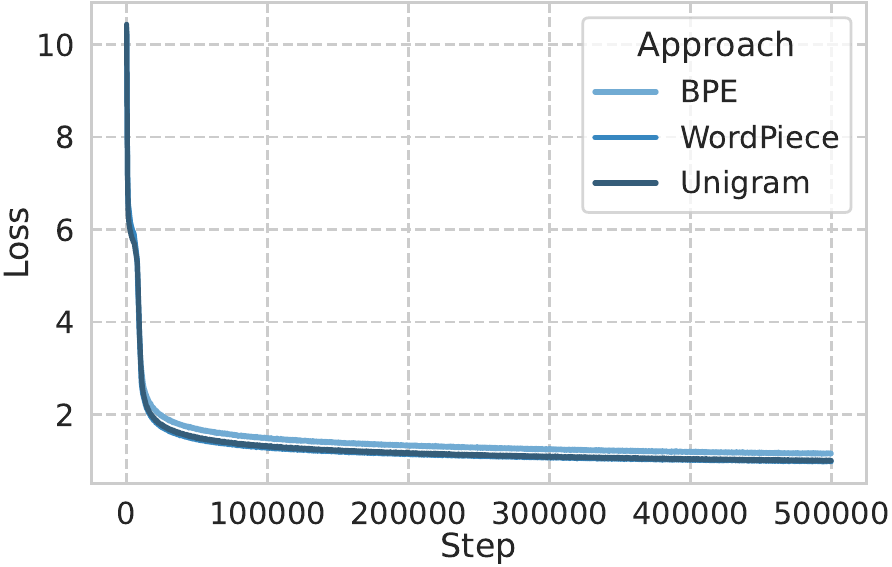}
    \subcaption{Vaporetto}
\end{minipage}
\hfill%
\begin{minipage}[b]{0.45\textwidth}
    \centering
    \includegraphics[width=\textwidth]{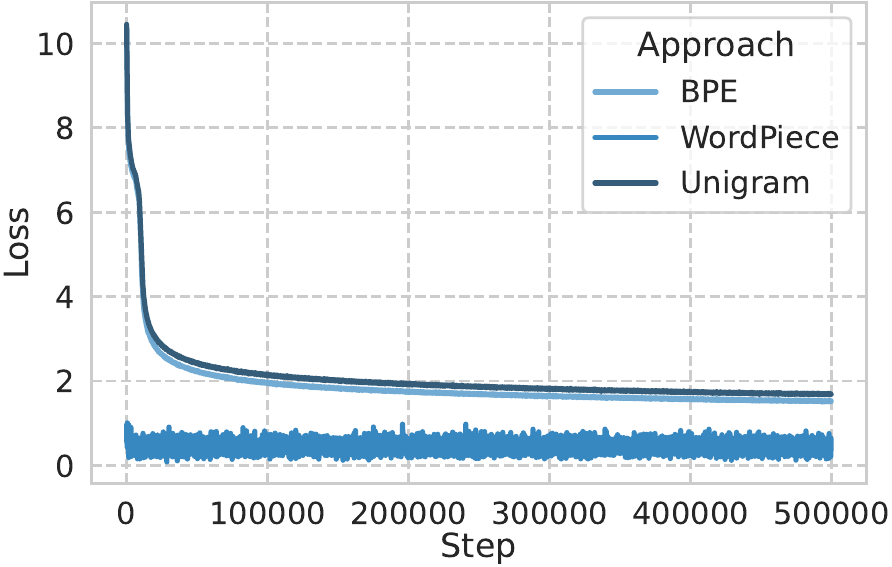}
    \subcaption{Nothing}
\end{minipage}
\caption{Pretraining loss curves}
\label{figure:loss}
\end{figure*}

Figure \ref{figure:loss} shows the pretraining loss curves for our models grouped by morphological analyzer.
We can see that WordPiece + Nothing was unstable in pretraining.

\end{document}